\documentclass[conference]{IEEEtran}
\IEEEoverridecommandlockouts

\usepackage{cite}
\usepackage{amsmath,amssymb,amsfonts}
\usepackage{algorithmic}
\usepackage{graphicx}
\usepackage{flushend}
\usepackage[numbers,sort&compress]{natbib}
\usepackage[capitalise]{cleveref}
\usepackage{textcomp}
\usepackage{xcolor,colortbl}
\usepackage{siunitx} 
\usepackage[table]{xcolor} 
\usepackage[top=54pt, left=54pt, right=54pt, bottom=54pt]{geometry}%
\def\BibTeX{{\rm B\kern-.05em{\sc i\kern-.025em b}\kern-.08em
    T\kern-.1667em\lower.2ex\hbox{E}\kern-.125emX}}

\definecolor{Gray}{gray}{0.75}
\definecolor{Grey}{gray}{0.92}

\newcolumntype{a}{>{\columncolor{Grey}}c}

\title{AMBER: A tether-deployable gripping crawler with compliant microspines for canopy manipulation}

\begin{document}


\author{P. A. Wigner$^{*1}$, L. Romanello$^{*2,3}$, A. Hammad$^{1}$, P. H. Nguyen$^{1}$, T. Lan$^{1}$, S. F. Armanini$^{4}$, B. B. Kocer$^{5}$, M. Kovac$^{2,3}$ \\

\thanks{ 
\par $^{1}$eAviation Laboratory, TUM School of Engineering and Design, Munich.
\par $^{2}$Laboratory of Sustainability Robotics, EMPA, Dübendorf, Switzerland.
\par $^{3}$ Laboratory of Sustainability Robotics, EPFL, Lausanne Switzerland.
\par $^{4}$Aerial Robotics Laboratory, Imperial College London.
\par $^{5}$Bristol University.}
}

\maketitle

\begin{abstract}
This paper presents an aerially deployable crawler designed for adaptive locomotion and manipulation within tree canopies. The system combines compliant microspine-based tracks, a dual-track rotary gripper, and an elastic tail, enabling secure attachment and stable traversal across branches of varying curvature and inclination. Experiments demonstrate reliable gripping up to 90° body roll and inclination, while effective climbing on branches inclined up to 67.5°, achieving a maximum speed of 0.55 body lengths per second on horizontal branches. The compliant tracks allow yaw steering of up to 10°, enhancing maneuverability on irregular surfaces. Power measurements show efficient operation with a dimensionless cost of transport over an order of magnitude lower than typical hovering power consumption in aerial robots. The crawler provides a robust, low-power platform for environmental sampling and in-canopy sensing. The aerial deployment is demonstrated at a conceptual and feasibility level, while full drone–crawler integration is left as future work.
\end{abstract}

\section{Introduction}
Forests host over 60,000 species, most of which are amphibians, birds, and mammals \cite{UNDP2024}, providing vital ecosystem services such as rainfall regulation, water protection, erosion control, and carbon storage \cite{WWF2024}. Tropical forests cover only 6\% of Earth's land but support nearly 80\% of known species and store vast carbon reserves \cite{leroy2019global, saatchi2011benchmark}. Deforestation and climate change are eroding this resilience, pushing ecosystems toward tipping points \cite{WWF2024}. Yet, despite their importance, the canopy ecosystems that sustain much of this biodiversity remain among the least accessible and studied environments.
Canopy studies traditionally rely on rope climbing, ladders, or cranes equipped with instruments such as weather stations, data loggers, and eDNA sensors \cite{Lowman1996ForestCanopies, Cevik2025eDNA}. These methods are labor-intensive, costly, and often disruptive, highlighting the need for efficient, low-impact alternatives \cite{hauer2017methods}. Epiphytes and branch-dwelling plants, which can comprise up to 50\% of local richness in some tropical forests \cite{Zotz2016}, play key roles in ecosystem function and serve as indicators of forest health, yet accessing them requires precise mobility along branches.

\begin{figure}[t!]
    \centering
    \includegraphics[width=1.0\linewidth]{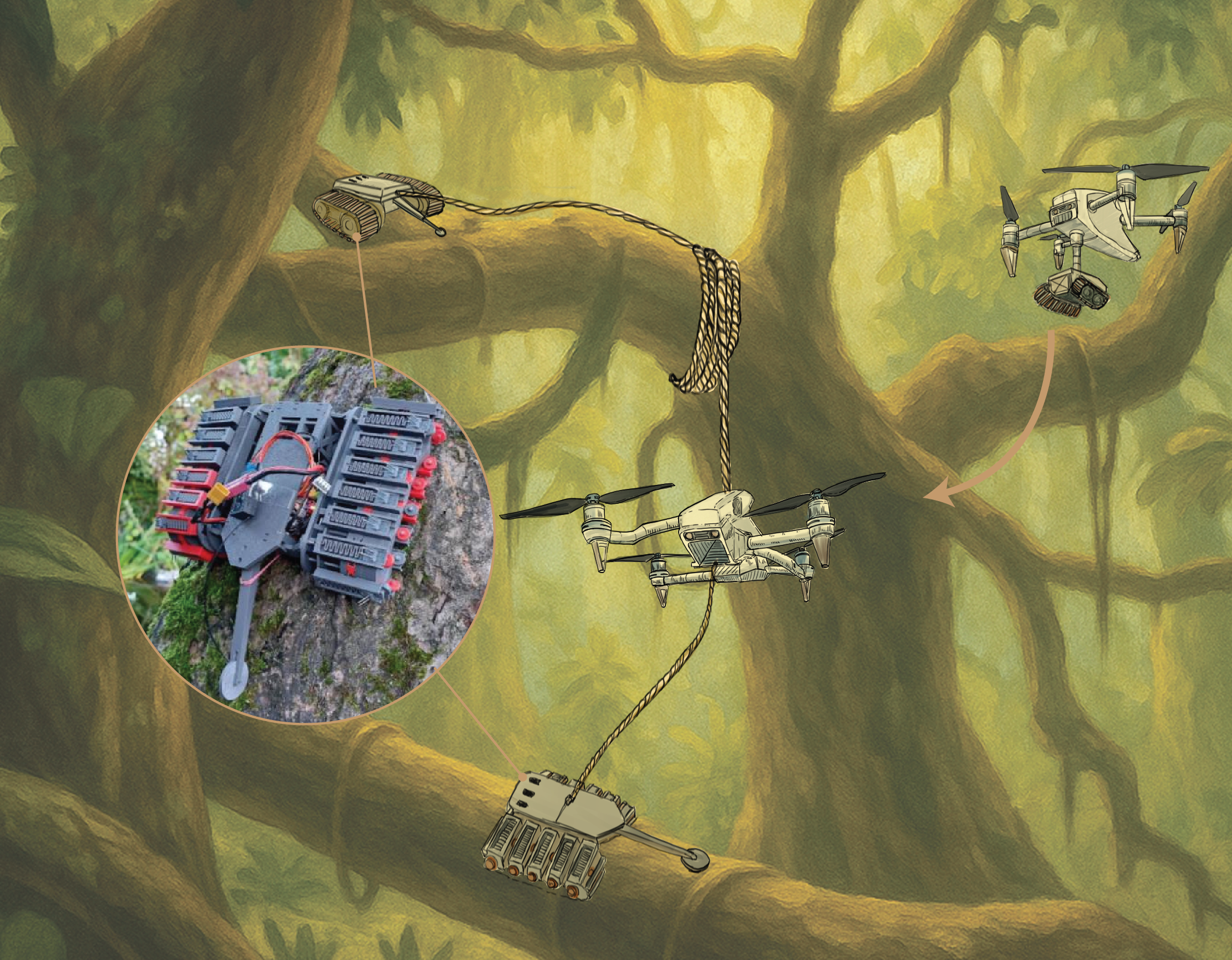}
    \caption{Conceptual mission illustration (not to scale). A drone deploys a tethered crawler onto a branch. Full aerial–crawler integration is future work.}
    \label{fig:concept}
\end{figure}

Robotic systems offer safer and more sustainable means of accessing the forest canopy. Drones are increasingly employed for ecological monitoring, yet their effectiveness remains constrained by propeller noise, limited flight endurance, and the difficulty of navigating dense vegetation with mid-sized aerial platforms \cite{UNEP2024}. Bioinspired perching mechanisms, including avian-inspired graspers \cite{hammad2025lightweight,roderick2021bird, doyle2011avian} and microspine-based systems \cite{kirchgeorg2021hedgehog, nguyen2019passively}, increase stability on branches but are constrained by geometry, size, and mounting beneath the drone. Tethered approaches allow flexible perching and integration of sampling pods \cite{hauf2023learning, 10609975, Romanello2026TreeSpider}, yet typically lack branch mobility.

Recent gripping robots show mobility on vertical surfaces, such as fixed-wing aircraft with microspines and suspension systems \cite{kim2006isprawl}, rotorcraft that perch and climb walls \cite{7797505}, and perching fixed-wings for tree trunks \cite{Askari2024CrashPerching}. However, these systems still lack the adaptability, continuous locomotion, and payload capacity required for canopy research.

To address these limitations, we present AMBER, a tether-deployed crawler for branch traversal and direct sampling. It features a dual-track mechanism with compliant microspines and a spring–damping suspension, enabling adaptive gripping on natural surfaces. As shown in \cref{fig:concept}, it can reach the canopy via aerial tether deployment. It combines energy-efficient perching with the mobility needed for in-canopy exploration and sensing, including payload transport along branches of varying diameters and inclinations and transitions onto trunks.  

\begin{figure*}[t!] \centering \includegraphics[width=0.9\linewidth]{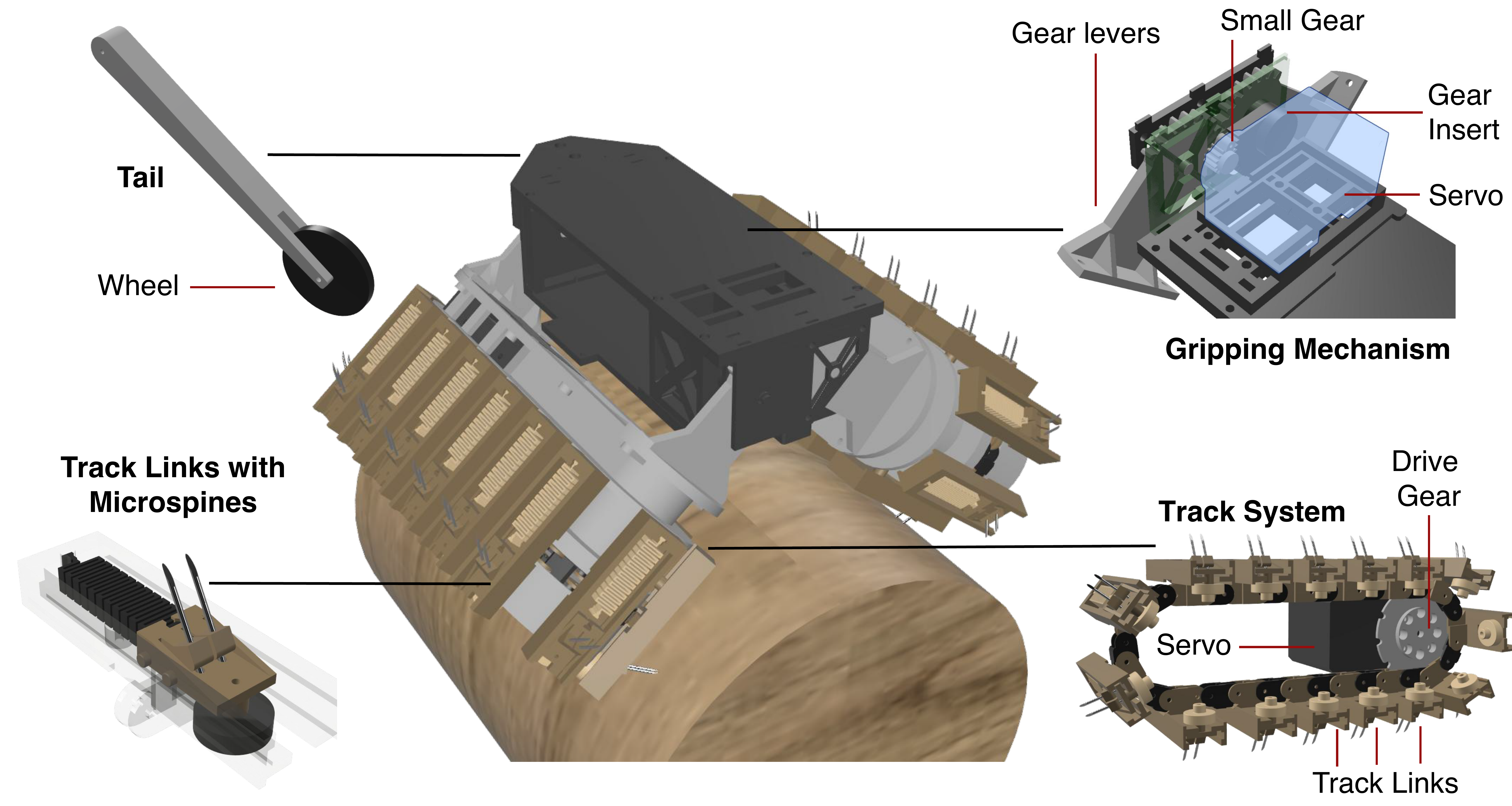} \caption{Crawler design with a focus on the gripping mechanism, the track system, the microspine element and the tail.} \label{fig:design} \end{figure*}

Existing climbing robots such as RiSE \cite{spenko_biologically_2008} and TreeBot \cite{lam_climbing_2011} demonstrate impressive vertical mobility using gecko-like adhesion and microspine grasping, yet they are either too heavy for aerial deployment or do not demonstrate complete crawling or gripping around the branch, which is useful for sampling epiphytes. Continuous-motion systems using wheels or tracks \cite{carpenter_rotary_2016, liu2015tbot, liu_novel_2019}, or trunk-enclosing designs \cite{ibrahim_design_2020}, remain confined to smooth or vertical surfaces. To overcome these limitations, AMBER introduces a novel framework for tree branch and trunk locomotion and manipulation, enabling rapid, aerial-deployable access for environmental sampling in complex canopy environments.
Aerial deployment enables rapid positioning of the crawler within the canopy, while the tether allows the drone to perch securely on a branch and provides safe recovery for the crawler, ensuring continued operation after a fall. The crawler is designed to accommodate lightweight sampling modules; however, this work focuses on locomotion and gripping performance. After sampling, the crawler is retrieved through the tether, providing a safe and cost-effective approach for in-canopy exploration and analysis. This capability also supports targeted ecological monitoring, such as detecting or limiting the spread of invasive pests and diseases, by positioning sensors on specific parts of the canopy.

The main contributions of this work are:

\begin{itemize}
\item \textbf{Tether-deployable canopy locomotion framework:} A lightweight crawler designed for deployment and recovery using an aerial tether, enabling safe positioning and retrieval in tree canopies.
\item \textbf{Novel adaptive crawler design:} A Dual-Track Rotary Grasper (DTRG) with compliant microspines and a spring–damping mechanism ensures robust attachment and smooth traversal on irregular branches. A passive wheeled tail enhances stability and grip during high-inclination climbing, extending mobility across complex canopy geometries.
\end{itemize}




\section{Methodologies}

Existing microspine-based track-type crawlers are mostly limited to flat surfaces, preventing traversal of sloped, uneven, or vertical terrain \cite{liu_track-type_2020, liu_novel_2019}. To overcome this, AMBER combines a track-type microspine system with an adaptive gripping mechanism, allowing the crawler to navigate branches from large trunks to smaller limbs efficiently. Passive gripping via a cam mechanism lets individual microspines move independently, ensuring secure perching even if some lose contact.

Aerial deployment allows rapid positioning in the canopy, while the tether lets the drone perch securely on branches and provides safe recovery for continued operation after a fall. When outfitted with a sampling mechanism, the crawler can traverse the tree structure and collect biological or environmental samples, supporting targeted ecological monitoring such as tracking or mitigating invasive pests and diseases. The design of AMBER is governed by three key constraints:\\

\begin{itemize}
\item \textbf{Dimensional:} The crawler must remain compatible with the size and mass range of aerial drones, ensuring integration or standalone operation. Compactness facilitates navigation between branches and grasping of different diameters, excluding bulky arm-based designs.
\item \textbf{Functional:} A continuous-track mechanism enables smooth, fast crawling, unlike legged systems (e.g., gecko- or inchworm-inspired) that rely on many actuators and exhibit discontinuous motion. Continuous locomotion reduces dynamic loads on tethered drones and allows steering through differential track control.
\item \textbf{Manufacturing:} The robot adopts a modular design for rapid prototyping and part exchange. Components are 3D-printed in PLA, with metal pins and screws reinforcing joints for strength and ease of replacement.
\end{itemize}


\subsection{Dual-Track mechanism (DTRG)} 

The DTRG mechanism enables the system to adapt and perch on branches of varying sizes and shapes. Its actuation range spans from a half-angle of $90^\circ$ (open) to $35^\circ$ (closed), with both sides gear-synchronized and driven by a single Dynamixel XL-330-M288-T motor, reducing mass and complexity. This configuration maintains the body tangential to the surface, enhancing clearance and minimizing collision risk.
The servo provides precise position and torque control, crucial for reliable attachment. Although nominally rated at 0.6~Nm (7.5~N at 80~mm radius), continuous operation required limiting current to 700~mA (0.250~Nm, $\sim$40\% of maximum torque). To compensate, a modular reduction stage was implemented: a $1\!:\!4$ gear ratio yielding up to 12.5~N gripping force within the 28~mm internal diameter, and an alternative $1\!:\!2$ stage with 7~mm minimum gear diameter.

\subsection{Crawler body}
The crawler body forms the central structural frame, serving as the load-bearing element for the gear levers, a tethered aerial drone, and the gripping servo motor. The system provides high rigidity and stability to ensure proper gear meshing and reliable track actuation while effectively restricting all non‑actuated degrees of freedom. All electronics—including the RC receiver, microcontroller, buck converter, motor, and 3S 850 mAh LiPo battery—are integrated within the body, defining its minimum dimensions of 54 by 37 mm based on the selected components. Gear and plate interfaces use bushings and tabs instead of bearings or screws to reduce thickness and maintain compactness, while additional slotted brackets carry shear loads and aid cooling. The body features an open layout with ventilation slots to prevent servo and electronics overheating, resulting in a total length of 120.1 mm that balances mechanical strength, compactness, and thermal management. The complete system weighs approximately 700 g, which, if carried by a commercial DJI F450 drone, would reduce flight duration by less than 50\% \cite{10609975}.

\subsection{Compliant microspines}

\begin{figure}[t!]
    \centering
    \includegraphics[width=0.9\linewidth]{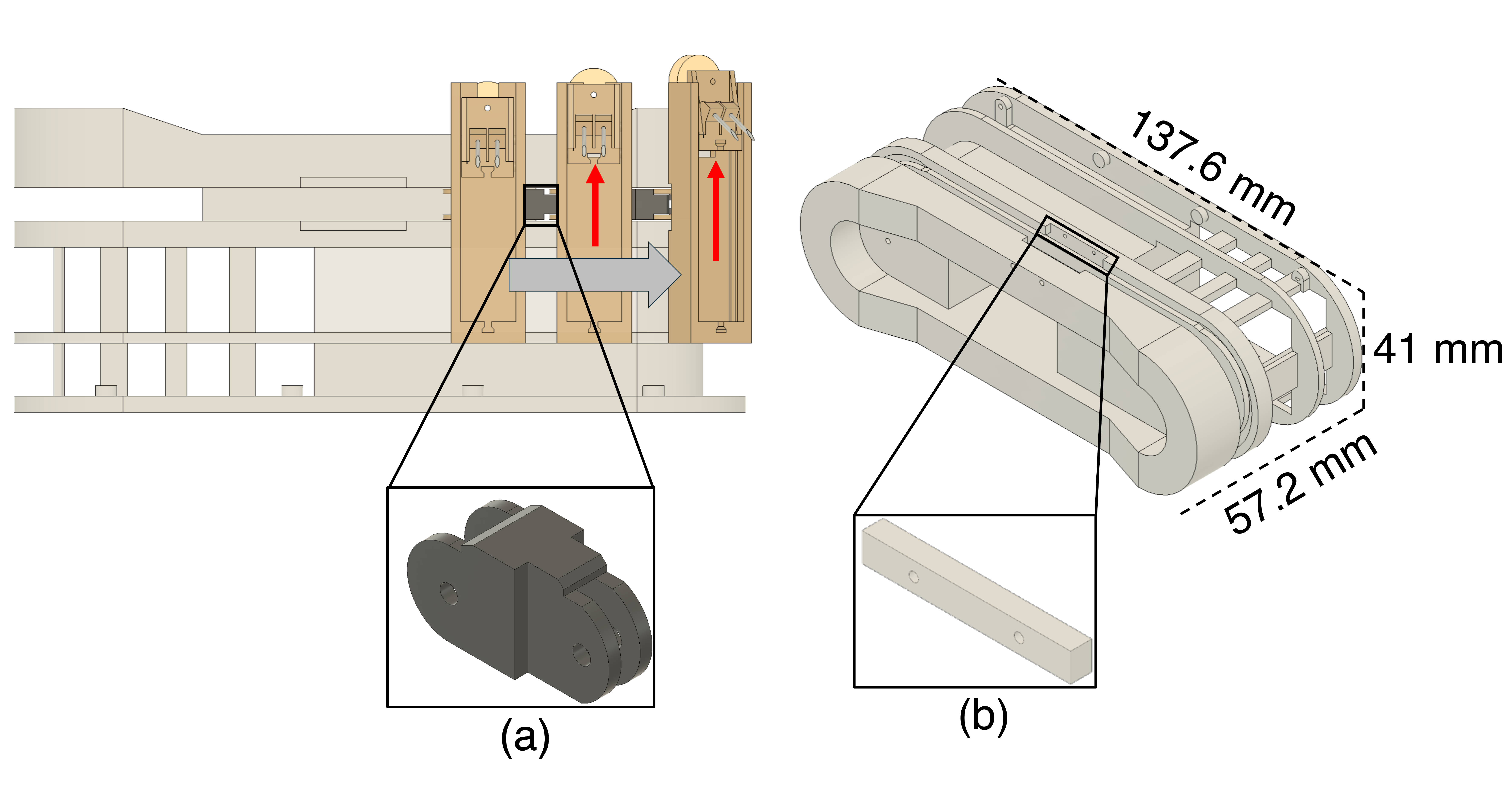}
    \caption{Track body components for spine carrier actuation: (a) intermediate chain link, (b) retainer closing the chain insertion slot during assembly, (c) spine carrier, and (d) track body.}
    \label{fig:microspine}
\end{figure}

\begin{figure}[t!]
    \centering
    \includegraphics[width=0.9\linewidth]{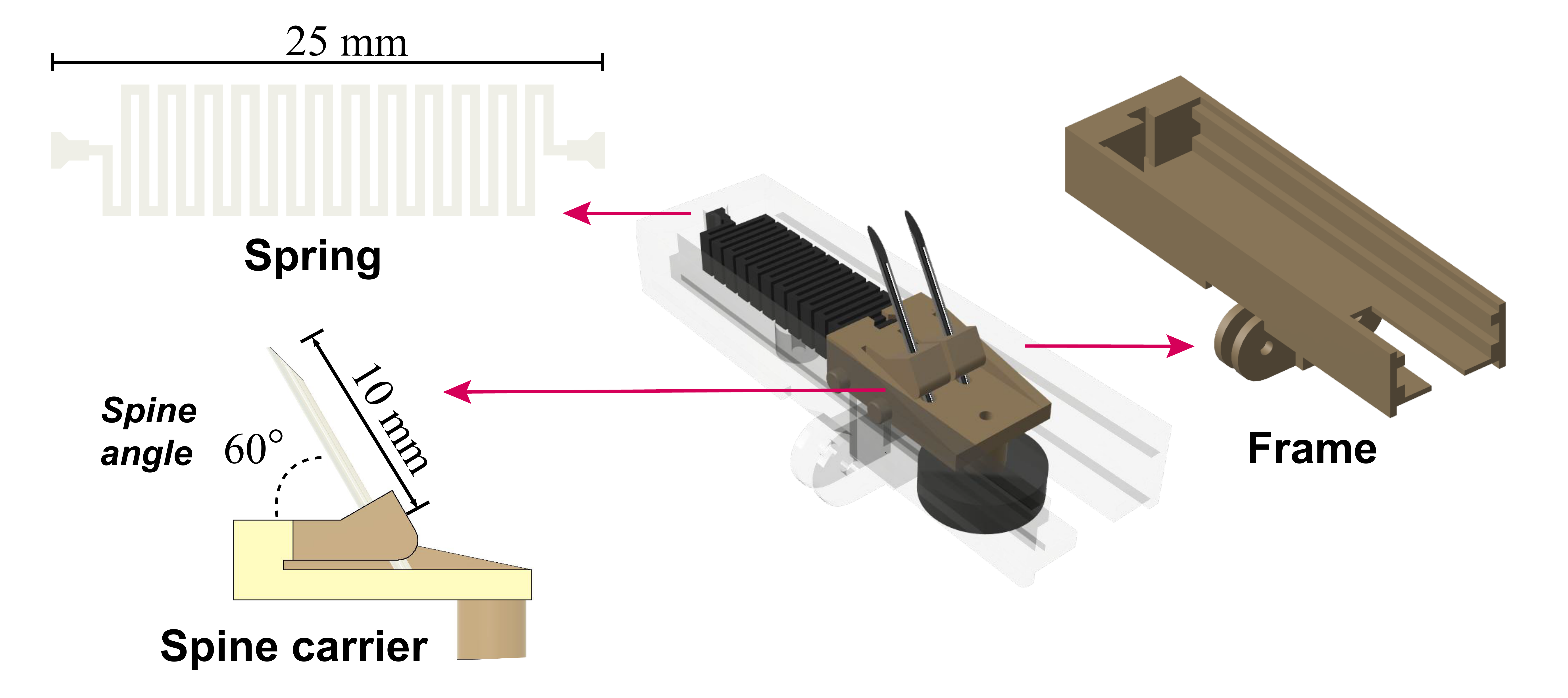}
    \caption{Compliant microspine design consisting of a frame, spring, and spine carrier, allowing lateral movement that facilitates crawler maneuverability.}
    \label{fig:microspine}
\end{figure}

Microspines were chosen for reliable attachment on irregular natural surfaces. Flexibly mounted, they engage asperities and share loads, preventing failure. Each track link has a lateral guide rail with a spine carrier connected to a 3D-printed PLA spring. Simulations determined optimal spring parameters of 0.7 mm thickness and 25 mm length (depicted in \cref{fig:microspine}), allowing 5 mm travel and 1 N force without plastic deformation.

The spine carrier permits lateral motion while preventing rotation, maintaining alignment and load distribution. Initial designs inspired by Liu et al. \cite{liu2015tbot} were too flexible for steep surfaces, and longer spines (10 mm) caused bending and misalignment. The revised stiffer carrier combines carrier and hollow spine flexibility to maintain compliance and grip, while also allowing yaw adjustment for crawler steering.

Surgical needles with ~10 µm tip radii were used to engage micro- and macroscopic asperities. Spine length and angle were experimentally evaluated on branches of 40, 85, 100, and 150 mm diameter (ten repetitions per configuration) under controlled conditions (21 °C, 200 mA, 5 V). Spine angles of 45° and 60° were compared.

Retention force generally increased with decreasing branch diameter, except for the rough 150 mm branch. Smaller branches allowed more wrapping and higher normal force. While 45° spines performed slightly better for some diameters, 60° spines offered more consistent grip and were selected for the final design (\cref{fig:microspine}).

\subsection{Track design}


To minimize weight for aerial integration, the body is manufactured mostly as thin-walled structures with material concentrated on the perimeter. Thicker elements in the upper section are printed partially hollow to reduce mass. Chain links are guided within a perimeter slot with an inner groove for the pins, constraining motion except along the track direction, as depicted in \cref{fig:microspine}. Pins are made from mild steel to withstand shear and bending loads, while smaller spacer links maintain consistent spacing between full links. 

The chain is driven by a gear directly mounted on the servo motor output shaft. All contact surfaces are lubricated with graphite to reduce friction and required power. The track body dimensions are constrained by the motor, drive gear, and spine actuation surfaces, resulting in a thickness of 41~mm and width of 57.2~mm. The distance between the horizontal centerline and the pin guide slot is 17~mm to avoid collisions.  

The actuation surfaces drive lateral spine movement, where bushings roll along the perimeter and translate the spine carriers outward, opening the gripper. As the links move along curves (red arrows on carrier in\cref{fig:microspine}) , the tension springs pull the carriers inward, forming the gripping motion. The springs exert 1~N at maximum extension, engaging three to five spine carriers at any time, allowing microspines to reliably grip surface asperities while distributing load evenly.

\subsection{Wheeled tail}

Long and mobile tails in arboreal animals improve stability when moving on narrow or precarious substrates by modulating body angular momentum and promoting balance during locomotion \cite{young2021tail}. Robotic studies have shown that adding a tail can significantly enhance locomotion performance \cite{siddall2021tails}. Building on this insight, we hypothesize that incorporating a tail into AMBER could improve branch-climbing ability by increasing grip and overall stability.

To minimize mass and control complexity, a passive tail-mounted wheel was selected, prioritizing robustness over optimal force generation. The tail, approximately 100 mm long, integrates a wheel with a tunable spring–damping system that reduces friction and improves stability during climbing. This compliance allows the wheel to passively regulate the applied pushing force, enabling secure and adaptive branch gripping and enhancing overall climbing performance. Inspired by prior climbing robots such as Tbot \cite{liu2015tbot} and soft-legged systems \cite{sadeghi2016softleg}, this design adapts established locomotion principles to a wheel-based tail architecture..

\subsection{Electronics \& Communication}
The crawler is equipped with an independent battery that powers the microcontroller (Arduino OpenRB 150) and, through a power converter, supplies energy to the three motors: two Dynamixel XL-430 units driving the tracks and one Dynamixel XL-330 operating the gripper. The battery also powers the RC receiver, which receives signals from the RC transmitter used to remotely control the robot’s speed, gripper angle, and yaw. These signals are processed by the microcontroller, which then outputs the corresponding commands to the motors.

\section{Experimental Results}

The robot and its features were validated through a series of comprehensive gripping and crawling tests designed to characterize the system’s complete operational performance across the three principal orientation axes. In this context, pitch refers to the inclination of the branch relative to gravity, roll denotes the robot’s rotation around the branch circumference, and yaw describes the steering misalignment between the crawler’s longitudinal axis and the branch. 

\subsection{Gripping}

\begin{figure}[t!]
    \centering
    \includegraphics[width=0.9\linewidth]{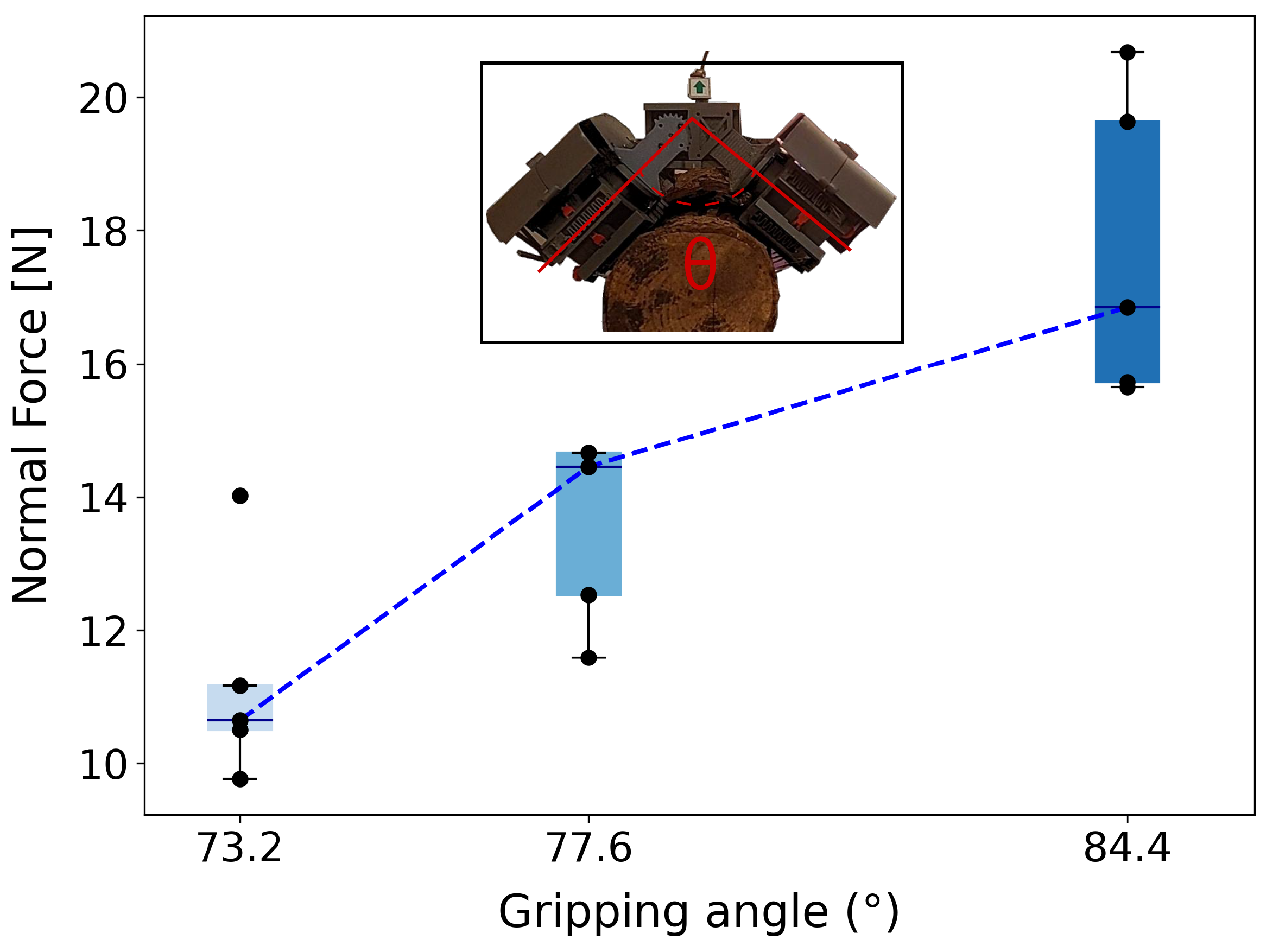}
    \caption{Normal pull-off force at different gripping angles.}
    \label{fig:gripping}
\end{figure}

For these tests, a branch with a diameter of 95 mm was used. The gripping angle was evaluated prior to testing the performance at various inclinations, yaw, and roll in order to maximize the normal force. The angle affects both the proximity of the crawler to the branch and the orientation at which the needles engage the branch. For this specific branch, the maximum pull-off force was recorded at \ang{84.4}, as shown in \cref{fig:gripping}, and this angle was therefore used for subsequent tests. A potential method to automatically adjust the angle based on branch characteristics would be to use a force sensor to detect when sufficient grip is achieved by the needles.

With the optimal gripping angle established, the crawler’s gripping performance was then evaluated on branches of varying inclinations under controlled conditions. Static tests altered the pitch (branch steepness), roll (angle around the branch), and yaw (steering angle) to assess gripping capability. Detachment forces were measured in the normal, tangential, and axial directions, with the crawler’s weight subtracted from normal-direction tests to isolate gripping force. Each configuration was tested five times, and the average of the maximum pull-off forces was computed.

\begin{figure*}[t!] \centering \includegraphics[width=\linewidth]{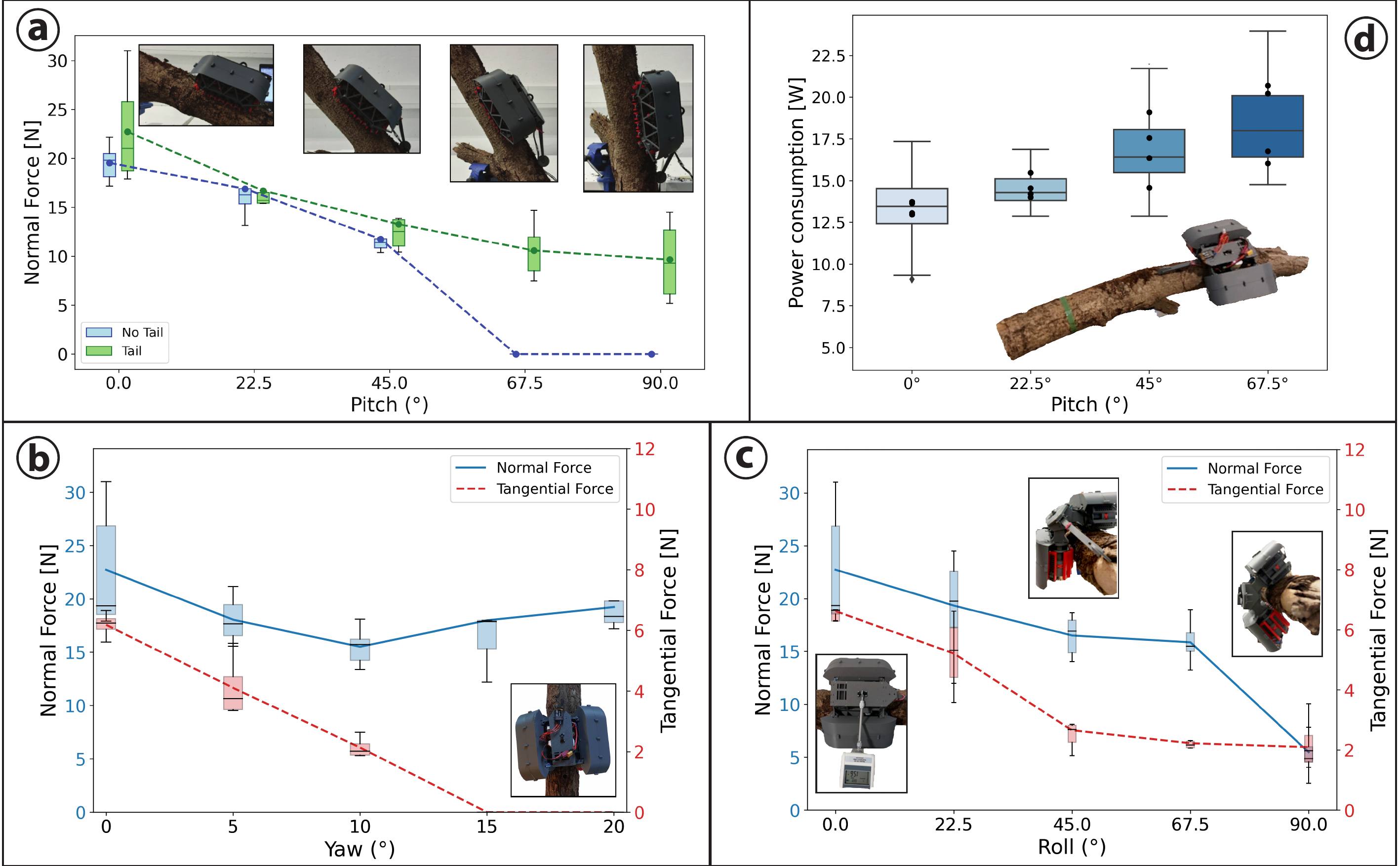} \caption{(a) Normal pull forces measured at various pitch angles for both the tail and no-tail configurations on a branch with a 95 mm diameter. (b) Pull off normal and tangential forces at different angles around the branch (Roll). (c) Pull off normal and tangential forces at Yaw angles up to 20°. (d) Power consumption while crawling at different inclinations. } \label{fig:tests_all} \end{figure*}


As shown in \cref{fig:tests_all}(a), maximum pull-off forces were recorded for configurations with and without the tail, confirming the tail’s importance on steep branches. Pull-off forces decreased with increasing pitch due to the weight-induced pitching moment. This is consistent with observations in arboreal quadrupeds, where tails act as stabilizing appendages that improve balance and mobility on narrow or inclined substrates \cite{young2021stabilizing}. The elastically loaded tail redistributed the load, increasing the maximum normal pull-off force by up to 15\% on inclinations below 45° (\cref{fig:tests_all}(a)) and enabling stable attachment on branches between 45° and 90°, which was not possible without the tail, while supporting payloads of up to 10 N.
The wheel relies on friction rather than grasping and is therefore sensitive to surface roughness and curvature.




With reference to \cref{fig:tests_all}(b), both normal and tangential forces decreased with increasing roll angle due to the center of mass shifting outward. Variations between repetitions were large, reflecting sensitivity to axial forces and slipping of the lower gripper. Despite this, the robot maintained a stable grip up to 90° roll. 

Finally, the system was tested at different steering angles (yaw) to assess maneuverability enabled by the compliant microspine system as shown in \cref{fig:tests_all}(c). At low angles, uneven force distribution caused some spines to detach first as the track element springs retracted, dynamically disrupting the gripper. The system then stabilized with force applied consistently at two contact points, slightly increasing pull-off force. For angles above 10°, only two spine carriers engaged, causing the crawler to tip due to limited contact points, making tests at higher angles infeasible.

\subsection{Crawling}

Crawling performance was evaluated on both straight beams and irregular branches, measuring maximum forward speed and grip reliability across different pitch angles. Maneuver reliability was defined as the ratio of successful trials to total trials, reflecting the crawler’s adaptability to variable surfaces.

\begin{figure}[t!]
    \centering
    \includegraphics[width=0.9\linewidth]{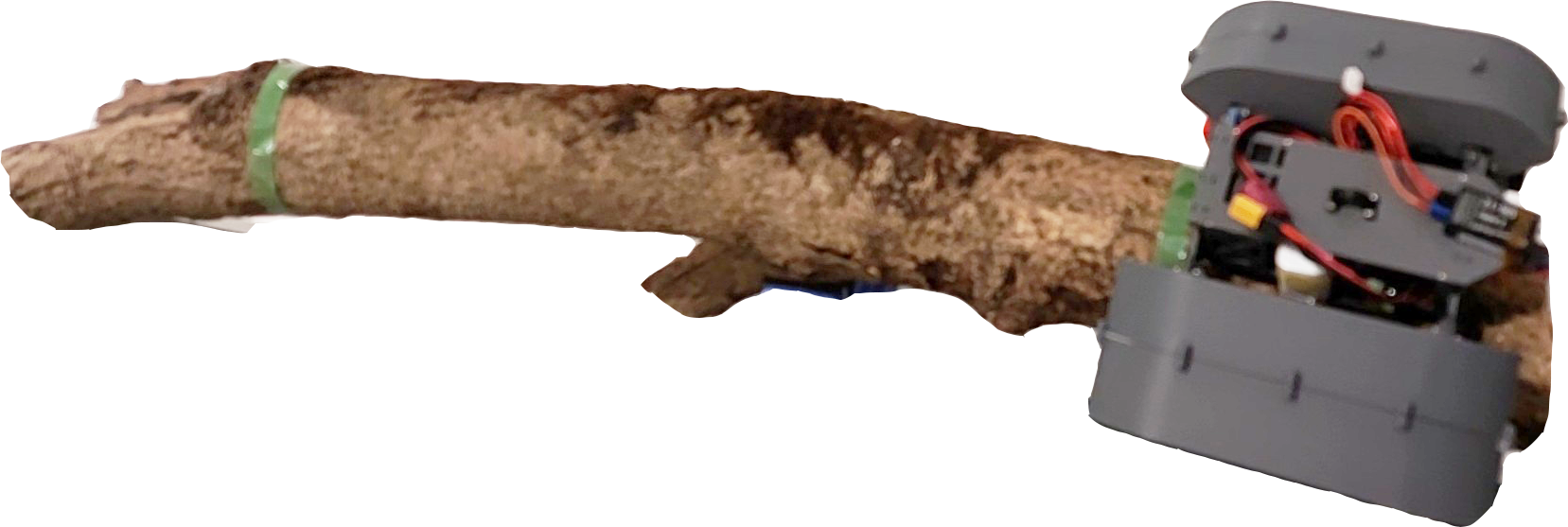}
    \vspace{0.1cm}
    \includegraphics[width=0.9\linewidth]{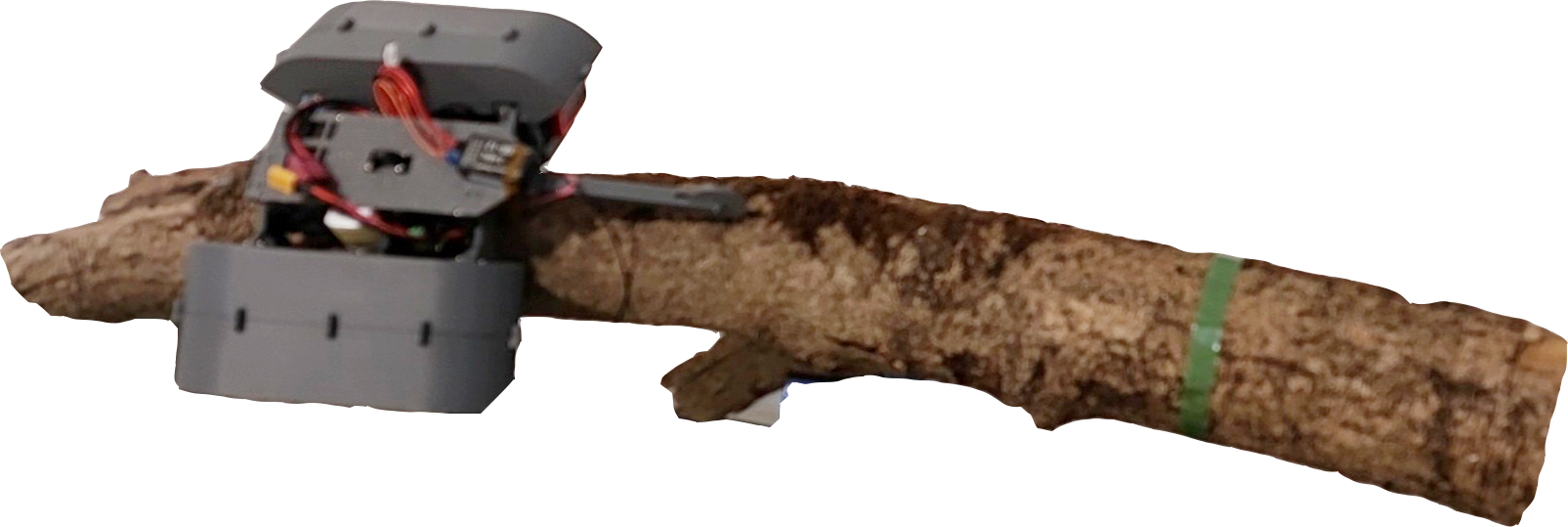}
    \caption{Crawling sequence on a horizontal branch: start (top) and end (bottom) positions of the manoeuvre, covering 50~cm at 0° inclination.}
    \label{fig:dynamic_crawling_sequence}
\end{figure}

\cref{fig:dynamic_crawling_sequence} shows the crawler moving along a horizontal branch, covering 50~cm. On a horizontal branch, the average speed was 8.33~cm/s (0.55 body lengths per second) with four out of five trials successful. Failures were primarily caused by branch curvature, highlighting the advantage of individually driven tracks. On inclines of \ang{22.5}, \ang{45}, and \ang{67.5}, average speeds decreased to 5.55, 4.61, and 4.16~cm/s, with corresponding success rates of 60\%, 50\%, and 33\%. These speeds are comparable to, or higher than, those of similar crawlers reported in the literature, which typically range from 1 to 4~cm/s \cite{spenko_biologically_2008, liu_track-type_2020, lam_climbing_2011}. Reduced climbing reliability at steeper inclinations was mainly due to insufficient tail support, causing pitching or rear detachment. Instabilities also occurred when high speeds coincided with surface irregularities; as the tail angle decreased, the elastic band shifted from its pivot, increasing the tail moment at an inappropriate time. An actively controlled tail could improve reliability, though at the cost of added weight and complexity.

\subsection{Power consumption \& Cost of Transport}

\begin{table}[h!]
\centering

\begin{tabular}{|l|S|S|S|S|}
\hline
\rowcolor{lightgray}
& {No-load} & {Perching} & {Crawling} & {Maximum} \\
\hline
Power [W] & 2.66 & 3.15 & 13.35 & 34.20 \\
\hline
\end{tabular}
\vspace{0.75em}
\caption{Power consumption in different operating conditions.}
\label{tab:Power_Consumption}
\end{table}

The crawler uses low-power servo actuation for gripping and propulsion. Power consumption was measured under different operating conditions. The corresponding Cost of Transport (COT) was computed as:

\begin{equation}
\text{COT} = \frac{P}{m g v} 
\end{equation}

\cref{tab:Power_Consumption,tab:COT_Consumption} summarize the results. No-load corresponds to the robot running without payload, with an average power of 9.85~W. Perching represents the robot holding position without movement, consuming only 3.15~W, nearly zero compared to continuous motion. Crawling on a horizontal branch required 13.35~W, while peak power under maximum load reached 34.2~W. The COT during horizontal crawling is 23.4, increasing to 64.6 at a \ang{67.5} incline, reflecting the higher energetic cost of climbing steeper surfaces.

\begin{table}[h!]
\centering
\begin{tabular}{|l|S|S|S|S|}
\hline
\rowcolor{lightgray}
Inclination & \ang{0} & \ang{22.5} & \ang{45} & \ang{67.5} \\
\hline
Speed [cm/s] & 8.33 & 5.55 & 4.61 & 4.16 \\
\hline
Power [W] & 13.35 & 14.55 & 16.90 & 18.46 \\
\hline
COT & 23.4 & 38.2 & 53.5 & 64.6 \\
\hline
Success Rate [\%] & 100 & 60 & 50 & 33 \\
\hline
\end{tabular}
\vspace{0.75em}
\caption{Power consumption, speed, and Cost of Transport (COT) while climbing different inclinations.}

\label{tab:COT_Consumption}
\end{table}

While aerial robots require continuous thrust, the crawler consumes only 3.15~W when perched and 13.35~W during horizontal crawling. A direct comparison with other tree crawlers is left for future work. 

This work presented a tether-deployable crawler designed for adaptive locomotion and manipulation within tree canopies. The system combines compliant microspine-based tracks, a dual-track rotary grasper, and an elastic tail, enabling secure attachment and traversal on branches of varying diameters and inclinations. Experiments demonstrated stable gripping on rolls up to 90° and climbing on branches up to 67.5° with a 33\% success rate. On horizontal branches, the robot achieved an average speed of 8.33 cm/s (0.55 body lengths/s), with yaw steering up to 10° enabled by the compliant microspine system. Power measurements indicated low energy requirements across modes, with Cost of Transport ranging from 23.4 on horizontal branches to 64.6 at 67.5°, showing efficient operation even on challenging terrain.

Key limitations remain. The current passive wheeled tail provides stability benefits but relies on frictional interaction rather than active force control; biological studies suggest tails enhance balance and stability in arboreal locomotion by modulating angular momentum on narrow supports, which motivates future active designs. Yaw steering is limited to low angles due to reduced contact at higher misalignment, and claims of full aerial deployment are conceptual rather than experimentally demonstrated in this work.

Future work will focus on enhancing tail-assisted stability through force‑ or geometry‑aware control of the gripping angle and active tail actuation to improve robustness on steeper inclinations. Additionally, integrating lightweight sensing and sampling tools will extend autonomous operation within forest canopies, enabling full drone–crawler mission execution in real-world field deployments.

\bibliographystyle{IEEEtran}
\bibliography{mybib}

\end{document}